\documentclass[letterpaper]{article} 
\usepackage{conference}  
\usepackage{times}  
\usepackage{helvet}  
\usepackage{courier}  
\usepackage[hyphens]{url}  
\usepackage{graphicx} 
\usepackage{natbib}  
\usepackage{caption} 
\usepackage{algorithm}
\usepackage{algorithmic}

\usepackage{newfloat}
\usepackage{listings}
\DeclareCaptionStyle{ruled}{labelfont=normalfont,labelsep=colon,strut=off} 
\floatstyle{ruled}
\newfloat{listing}{tb}{lst}{}
\floatname{listing}{Listing}
\usepackage{graphicx}
\usepackage{amsmath}
\usepackage{amssymb}
\usepackage{booktabs}

\usepackage{url}            
\usepackage{booktabs}       
\usepackage{amsfonts}       
\usepackage{nicefrac}       
\usepackage{algorithm}
\usepackage{algorithmic}
\usepackage{graphicx}
\usepackage{textcomp}
\usepackage{xcolor}
\usepackage{subcaption}

\usepackage{color}
\usepackage{epsfig}
\usepackage{graphicx}

\usepackage{adjustbox}
\usepackage{array}
\usepackage{booktabs}
\usepackage{multirow}

\usepackage{changepage}
\usepackage{extramarks}
\usepackage{fancyhdr}
\usepackage{lastpage}
\usepackage{setspace}
\usepackage{soul}
\usepackage{xspace}

\usepackage{comment}
\usepackage{enumitem}
\usepackage{chngcntr}

\definecolor{BlueViolet}{rgb}{0.26, 0.21, 0.58}
\definecolor{AmericanRose}{rgb}{0.69, 0.05, 0.05}
\definecolor{green(pigment)}{rgb}{0.0, 0.65, 0.31}

\renewcommand{\underline}[1]{#1}

\title{Synthetic Data Can Also Teach: \\ 
Synthesizing Effective Data for Unsupervised Visual Representation Learning}
\author{
    Yawen Wu\textsuperscript{\rm 1}\thanks{Corresponding author.}, 
    Zhepeng Wang\textsuperscript{\rm 2},
    Dewen Zeng\textsuperscript{\rm 3},
    Yiyu Shi\textsuperscript{\rm 3},
    Jingtong Hu\textsuperscript{\rm 1},
}
\affiliations{
    \textsuperscript{\rm 1}University of Pittsburgh, 
    \textsuperscript{\rm 2}George Mason University, 
    \textsuperscript{\rm 3}University of Notre Dame, 

    \{yawen.wu, jthu\}@pitt.edu, 
    zwang48@gmu.edu,
    \{dzeng2, yshi4\}@nd.edu
}

\usepackage{bibentry}

\begin{document}

\maketitle

\begin{abstract}
Contrastive learning (CL), a self-supervised learning approach, can effectively learn visual representations from unlabeled data. 
Given the CL training data, generative models can be trained to generate synthetic data to supplement the real data. Using both synthetic and real data for CL training has the potential to improve the quality of learned representations.
However, synthetic data usually has lower quality than real data, and using synthetic data may not improve CL compared with using real data.
To tackle this problem, we propose a \texttt{data generation} framework with two methods to improve CL training by joint sample generation and contrastive learning.
The first approach generates hard samples for the main model.
The generator is jointly learned with the main model to dynamically customize hard samples based on the training state of the main model.
Besides, a pair of data generators are proposed to generate similar but distinct samples as positive pairs. In joint learning, the hardness of a positive pair is progressively increased by decreasing their similarity.
Experimental results on multiple datasets show superior accuracy and data efficiency of the proposed data generation methods applied to CL.
For example, about 4.0\%, 3.5\%, and 2.6\% accuracy improvements for linear classification are observed on ImageNet-100, CIFAR-100, and CIFAR-10, respectively.
Besides, up to 2$\times$ data efficiency for linear classification and up to 5$\times$ data efficiency for transfer learning are achieved.
\end{abstract}

\section{Introduction}
Contrastive learning (CL), a highly effective self-supervised learning approach \cite{chen2020simple,he2020momentum}, has shown great promise to learn visual representations from unlabeled data.
CL performs a proxy task of instance discrimination to learn data representations without requiring labels, leading to well-clustered and transferable representations for downstream tasks.
In the proxy task, the representations of two transformations of one image (a positive pair) are pulled close to each other and pushed away from the representations of other samples (negatives), by which high-quality representations are learned \cite{kalantidis2020hard}.

Most recent CL works focus on developing CL training methods such as constructing contrastive losses for improving the learned representations \cite{caron2020unsupervised,zbontar2021barlow,Chen2021ExploringSS,grill2020bootstrap}, 
while what \texttt{data} to use for CL training remains largely unexplored.
Training with more data has the potential to improve existing CL methods.
This is because CL learns by comparing different pairs of samples (different from supervised learning which learns from every single sample and its label). By providing more data, more diverse pairs of data can be fed into CL models to learn better visual representations.
However, collecting more data to supplement existing data is usually very expensive.
For example, the ImageNet dataset \cite{ILSVRC15} widely used in CL has more than a million images of natural scenes. 
Collecting such large-scale datasets requires years of considerable human effort \cite{zhao2020differentiable}.
While it seems effortless to acquire these pre-collected datasets by simply downloading, collecting more data of natural scenes to supplement existing data such as doubling the number of samples requires another years of effort, which is very expensive or even prohibitive.
Therefore, without collecting more data, it is crucial to extract as much information as possible from existing data.

\begin{figure}[t!]
	\centering
	\includegraphics[width=1.0\linewidth]{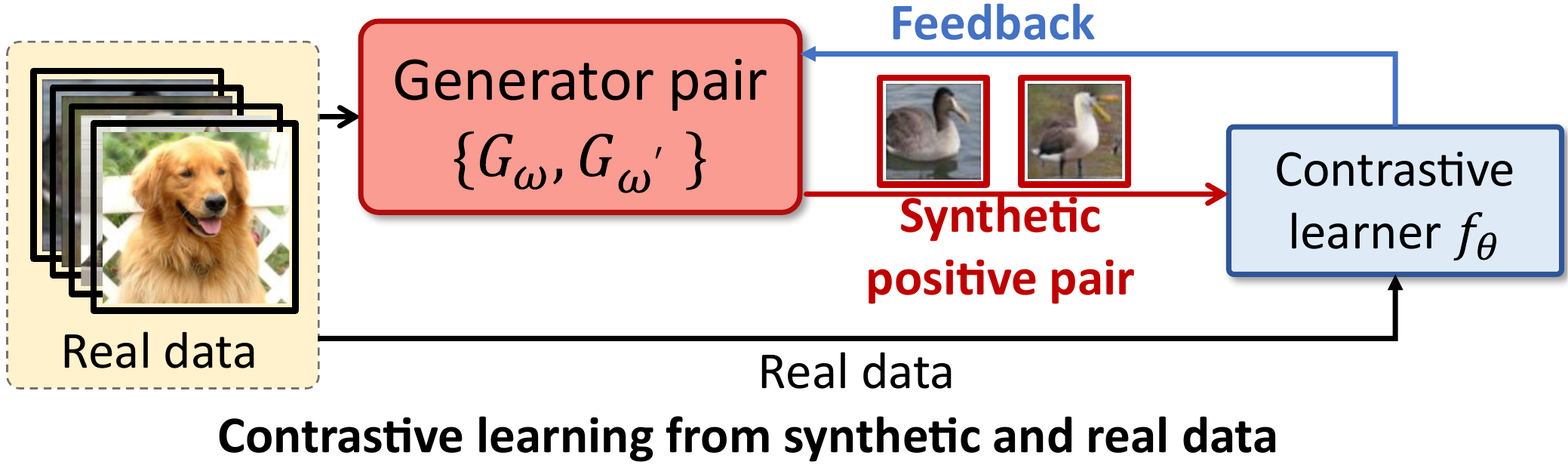}
	\caption{
		In this work, we study how to use generative models to synthesize additional data, such that using the synthetic data can improve contrastive learning (CL). 
		A pair of generators is employed to generate pairs of data as a positive pair for CL.
		The generators are jointly optimized with the contrastive learner to generate data beneficial to the learner.
		The proposed joint optimization is needed to improve CL.
		Without joint optimization, simply synthesizing data will degrade the performance of CL.
		}
	\label{fig:overview_compare}
\end{figure}

Our work therefore investigates a problem that has received little prior emphasis: given an unlabeled training set, can we 
generate synthetic data based on this dataset, such that using both synthetic and real training data can improve existing CL methods than only using real data?

Towards this goal, we propose a data generation framework to generate effective data for CL learning based on given training data. As shown in Fig. \ref{fig:overview_compare},
the data generation and CL model are jointly optimized by using the given training data, and no additional data needs to be collected.
The framework consists of two approaches. 
The first approach generates hard samples for the main contrastive model. 
The generated samples dynamically adapt to the training state of the main contrastive model by tracking the contrastive loss, rather than fixed throughout the whole training process.
With the progressively growing knowledge of the main model, the generated samples also become harder to encourage the main model to learn better representations.
The hard samples adversarially explore the weakness of the main model, which forces it to learn discriminative features
and improves the generalization performance.

The second approach generates two similar but distinct images as hard positive pairs.
Existing CL frameworks form a positive pair by applying two data transformations (e.g. color distortions) to one image to generate two transformed images.
While the two transformed images look different, they still share the same identity since they originate from one image. Only clustering these positive pairs will limit the quality of learned representation since other similar objects are not considered in clustering.
We form hard positive pairs by generating two images of distinct identities but similar objects without using labels, which is achieved by using a generator and its slowly evolving version.
In joint learning, the positive pair becomes harder by decreasing their similarity.
The main model has to learn to cluster hard positives when minimizing the contrastive loss.
By pulling the representations of similar but distinct (hard) objects together, better clustering of the representation space can be learned \cite{khosla2020supervised,wu2022decentralized}.
With better representations, the performance of downstream tasks will also be improved.

In summary, the main contributions of the paper include:

\begin{itemize}[itemsep=0pt,topsep=0pt,leftmargin=12pt]
	\item \textbf{Data generation framework for contrastive learning.}
	We propose a data generation framework with two approaches to synthesize effective training data for contrastive learning by jointly optimizing data generation and contrastive learning.
	The first approach generates hard samples
	and the second approach generates hard positive pairs without using labels.
	By applying this framework to existing CL methods, better representation can be learned.
	\item \textbf{Dynamic hard samples generation by tracking contrastive loss.} 
	We propose an approach to generate hard samples
	by dynamically tracking the training state of the main model. 
	In the joint learning process, hard samples are customized on the fly to the progressive knowledge of the main model, which are fed into the main model to constantly encourage the main model to learn better representations.
	\item \textbf{Hard positive pair generation without using labels.} 
	We propose an approach to further generate hard positive pairs without leveraging labels. The generator and its slowly evolving version generate a pair of similar but distinct objects as a positive pair. The hardness of a positive pair is further increased by decreasing their similarity in joint learning. 
    By learning from hard positive pairs, similar objects are well-clustered for better representations.
\end{itemize}

\section{Background and Related Work}

\textbf{Revisiting Contrastive Learning.}
Contrastive learning is a self-supervised approach to learning an encoder for extracting data representations from unlabeled images by performing a proxy task of instance discrimination \cite{chen2020simple,he2020momentum,wu2018unsupervised,tang2022contrastive}.

Our work in this paper is built upon SimCLR \cite{chen2020simple}, which is a simple yet powerful contrastive learning approach for unsupervised representation learning.
For an input image $x$, its representation vector $v$ is obtained by $v=f(x, \theta),\ z\in \mathbb{R}^d$, 
where $f(\cdot, \theta)$ is the encoder with parameters $\theta$.
In the training process of CL, a raw batch of $N$ samples $\{x_k\}_{k=1...N}$ are first randomly sampled from the dataset. Then for each raw sample $x_k$, two transformations ($t \sim \mathcal{T}$ and $t^{\prime} \sim \mathcal{T}$) sampled from a family of augmentations $\mathcal{T}$ (e.g. cropping, color distortion, etc.) are applied to $x_k$ to generate two transformed samples (a.k.a. two views). $\tilde{x}_{2k-1}=t(x_k)$ and $\tilde{x}_{2k}=t^{\prime}(x_k)$ are a positive pair, and all of the generated pairs form the batch for training $\{\tilde{x}_l\}_{l=1...2N}$, consisting of $2N$ samples \cite{khosla2020supervised}.
In the remainder of this paper, we will refer to the set of $N$ samples as a \textit{raw batch} and the set of $2N$ transformed samples as a \textit{multiviewed batch}.

In the multiviewed batch, let $i \in I=\{1...2N\}$ be the index of a transformed sample, 
and let $j(i)$ be the index of the transformed sample originating from the same raw sample as $i$. 
The contrastive loss is as follows.
\begin{equation}\label{equ:cl_loss}
\mathcal{L}_{CL} = \sum_{i\in I} \mathcal{L}_{i} = - \sum_{i\in I} \log \frac{\exp{(v_i \cdot v_{j(i)} / \tau)}}{\sum_{a \in A(i)}{\exp (v_i \cdot v_a / \tau)}}.
\end{equation}
where $v_i=f(\tilde{x}_i, \theta)$, the operator $\cdot$ is the inner product to compute the cosine similarity of two vectors, $\tau$ is the temperature. 
The index $i$ is the anchor. 
$A(i)=I\backslash\{i\}$ is the set of indices excluding $i$. 
For each anchor $i$, there is one positive and $2N-2$ negatives.
The index $j(i)$ is the positive to $i$ (i.e. a positive pair $(i,j(i))$), 
while other $2N-2$ indices $\{ k\in A(i) \backslash \ \{j(i)\} \}$ are the negatives.

Existing works focus on developing contrastive learning methods, without considering what data to use for CL.
Different from these works, we investigate CL from the data perspective.
That is, given the training data, how to generate more effective data for CL without collecting more data.

\begin{figure*}[ht]
	\centering
	\includegraphics[width=0.9\linewidth]{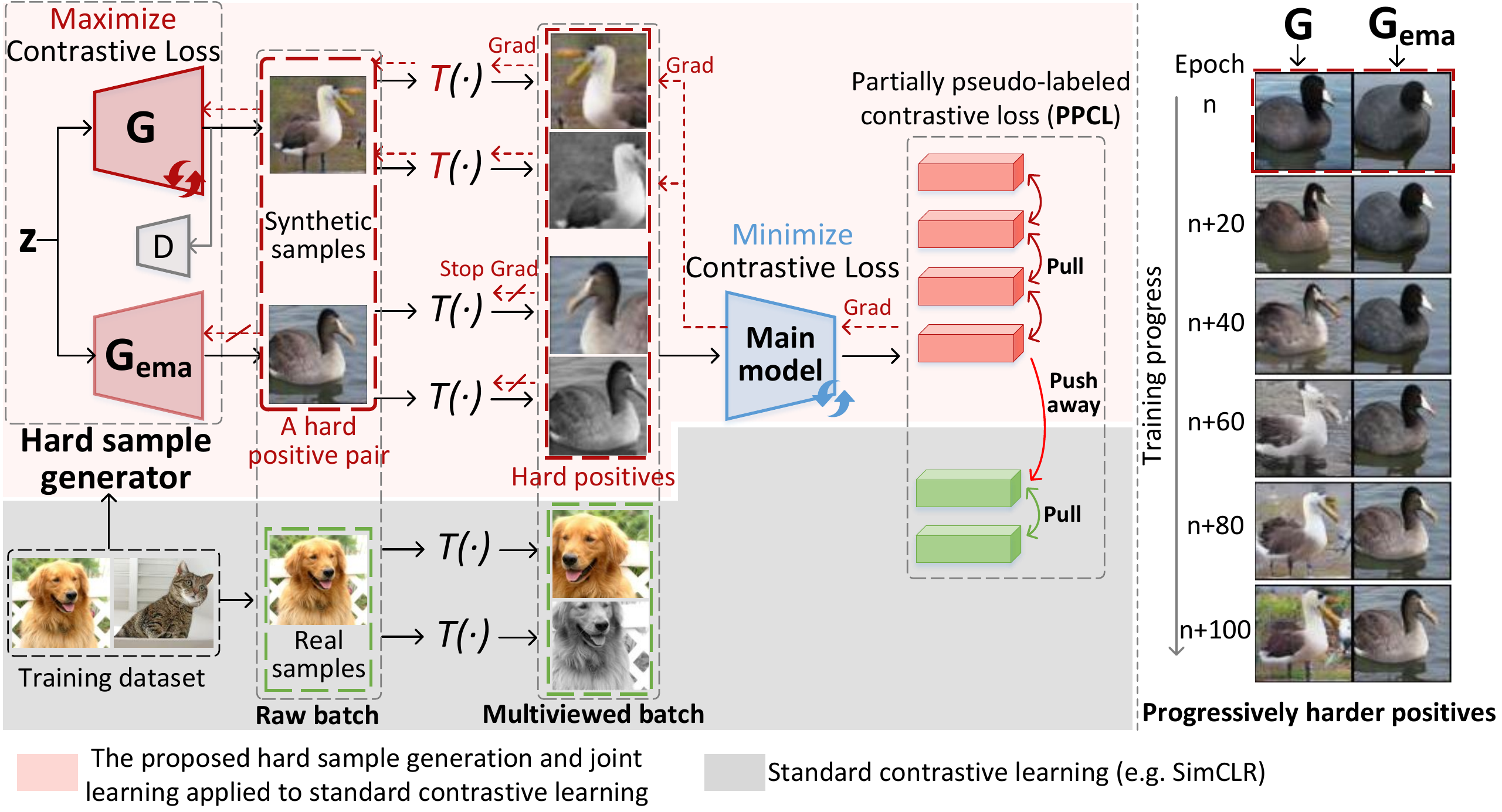}
	\caption{(Left) Generation of hard samples and hard positive pairs, and the joint learning of generator and the main contrastive model.
	We generate two similar but distinct raw samples, and use two views of each raw sample (four views in total) as positives, which are then fed into standard CL frameworks (e.g. SimCLR) for learning.
	No additional training data is used and no labels are used in the entire training pipeline.
	(Right) By joint learning, the generated positive pair becomes progressively harder by tracking the training state of the main model. These hard positive pairs help the main model cluster distinct yet similar objects for better representations.}
	\label{fig:diff_hard_sample_generation}
\end{figure*}

\textbf{Adversarial Samples for Improving Accuracy and Robustness of CL.}
To improve the quality of learned representations, 
adversarial attacks can be used to create additional training samples by adding pixel-level perturbations to clean samples \cite{ho2020contrastive}. 
Feature-selection based methods \cite{bao2020fast,bao2022accelerated} also have the potential to create adversarial samples by masking pixel-level features.
While adversarial samples are more challenging and can generate a higher loss than the original samples, 
the perturbed samples still have the same identities as the original ones, 
which provide limited additional information for learning. 
Besides, training with adversarial samples is originally designed for the robustness of models against attacks, instead of improving model performance on clean samples \cite{kurakin2016adversarial,madry2017towards,goodfellow2014explaining}. 
As a result, only marginal improvement \cite{ho2020contrastive} or even degraded performance \cite{kim2020adversarial,jiang2020robust} of the learned CL model is observed.
Different from these works, we generate whole images directly, instead of adding pixel-level noises to the existing images, which are more informative for improving the learned representations of CL.

\textbf{GAN for Data Augmentation.}
\cite{zhang2019dada,bowles2018gan,antoniou2017data,perez2017effectiveness} employ supervised class conditional GAN to augment the training data to improve classification performance. However, these works require fully labeled datasets for training GAN.
Since labels are not available in CL, the quality of images from GAN will greatly degrade \cite{miyato2018spectral,zhao2020differentiable} and the performance of trained CL model also degrades.
Besides, either GAN and the main model are isolated and the generated data are not adapted to the training state of the main model \cite{zhang2019dada,bowles2018gan,antoniou2017data}, or both GAN and the classification model aim to minimize the classification loss \cite{perez2017effectiveness}.
Different from these works, our methods do not rely on labels. 
Besides, 
the generator and main model are jointly learned,
in the way that the generator aims to maximize the CL loss while the CL model aims to minimize the CL loss.
Also, we generate hard positive pairs for unsupervised representation learning by CL, which is unexplored in these works on conventional supervised learning.

\section{Method}

\begin{figure}[H]
	\centering
	\begin{subfigure}[b]{0.495\linewidth}
		\includegraphics[width=\linewidth]{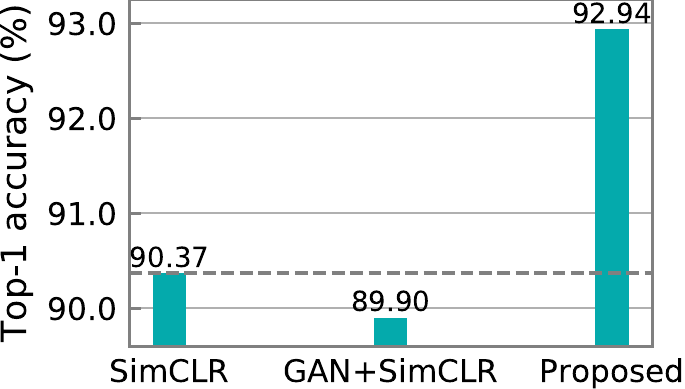}
		\caption{CIFAR-10.}
		\label{fig:}
	\end{subfigure}
	\begin{subfigure}[b]{0.495\linewidth}
		\includegraphics[width=\linewidth]{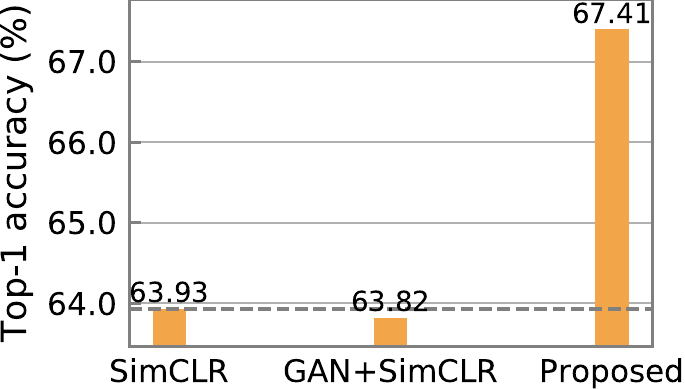}
		\caption{CIFAR-100.}
		\label{fig:}
	\end{subfigure}
	\caption{\textbf{SimCLR's performance is not improved when additional training data is simply provided by GAN due to the low quality of the synthetic data.} Different from this, with the proposed hard sample generation and joint learning, we are able to dramatically improve SimCLR's performance by +2.6\% and 3.5\% top-1 accuracy on CIFAR-10 and CIFAR-100, respectively.}
	\label{fig:motivation}
\end{figure}

We propose a data generation framework to generate effective training data for contrastive learning.
The framework generates individually hard samples and hard positive pairs without using labels, such that better representations can be learned from the unlabeled training data. 
We illustrate our data generation method by applying it to a typical CL framework SimCLR \cite{chen2020simple} as shown in Fig. \ref{fig:diff_hard_sample_generation}. However, our method can also be applied to other existing CL methods.

\textbf{Challenge: Low quality of synthetic samples degrades contrastive learning.}
Simply using a GAN pre-trained on the same unlabeled dataset as contrastive learning to provide additional synthetic data to the main model cannot effectively improve and even degrades the learned representation as shown in Fig. \ref{fig:motivation}.
This is because the synthetic data has intrinsically lower quality than the real data \cite{brock2018large}
even if labels were available to train a class-conditional generator. When the dataset is unlabeled and the generator is trained in a non class-conditional way, the quality of synthetic data becomes worse \cite{zhao2020differentiable,miyato2018spectral}, which degrades the performance of the main model or only provides marginal benefits.

To solve this problem, instead of using the standalone generator and contrastive main model, we jointly optimize them by formulating a Min-Max game such that they compete with each other.
As shown in Fig. \ref{fig:diff_hard_sample_generation}, there are two major components: the hard sample generator (red) and the main contrastive model (blue), which are jointly optimized.
Joint learning effectively uses the available unlabeled training data, and no additional training data or labels are used.

Algorithm~\ref{alg:diffhard} shows the process of joint contrastive learning with hard sample and hard positive pair generation. 
To be a clean self-supervised method, labels are never used in the entire generator pre-training and the joint contrastive learning process. 
We formulate the joint learning process as a Min-Max game, where the generator maximizes the contrastive loss by generating hard samples, while the main model minimizes the contrastive loss by learning representations from both the generated and real samples. The sample generator consists of two components, the generator $G$ and its slowly-evolving version $G_{\text{ema}}$.
$G$ and $G_{\text{ema}}$ are first pre-trained with a discriminator $D$ on the unlabeled dataset by using the unconditional GAN objective \cite{brock2018large} to generate data following the real data distribution.
Then the joint hard sample generation and contrastive learning start, which has 3 steps. 
First, $G$ generates individually hard samples, and $G$ and $G_{\text{ema}}$ collaboratively generate pairs of hard positives.
The hard samples, hard positives, and real samples from the dataset form a batch and are fed into the main model to compute the contrastive loss.
After that, main model $f$ is updated to minimize the contrastive loss.
Finally, $G$ is updated to maximize the contrastive loss to generate harder samples for the main model based on the current training state of the main model. Momentum update is applied to $G_{\text{ema}}$ to follow $G$. Meanwhile, we are using $D$ to force $G$ to generate meaningful data following real data distributions. In this way, the generator and the main model are jointly optimized, such that we can generate progressively harder samples and adapt to the training progress of the main model as shown in Fig. \ref{fig:diff_hard_sample_generation} (Right). The details of each step will be discussed in the 
following subsections.

\begin{figure}[!htb]
\begin{minipage}{1.0\linewidth}
\begin{algorithm}[H]
	\caption{Contrastive Visual Representation Learning with Synthetic and Real Data}
	\label{alg:diffhard}
	\begin{algorithmic}\itemindent=-10pt
		\STATE {\bfseries Input:} Unlabeled dataset $\mathcal{S}$, main model $f(\cdot, \theta)$, generator pair  $\{G(\cdot, w),G_{\text{ema}}(\cdot, w_{\text{ema}})\}$, discriminator $D(\cdot, \phi)$.
		\STATE {\bfseries Output:} Trained main model $f(\cdot, \theta)$.
		\STATE Pre-train $G$, $G_{\text{ema}}$ and $D$ 
        by the unconditional GAN objective on the same unlabeled training dataset $\mathcal{S}$ as CL.
		\STATE \textcolor{BlueViolet}{\# Joint learning}
		\FOR{iteration $t=1$ {\bfseries to} $\mathrm{T}$}
		\STATE \textcolor{BlueViolet}{\# Step 1: Forming a data batch with synthetic and real samples}
		\STATE Sample generated batch $B_{\text{gen}}$ following Eq.(\ref{equ:generate_pair}) and real batch $B_{\text{real}}$ from $\mathcal{S}$.
		\STATE Form multiviewed batch $B_{\text{mv}}$ from $\{B_{\text{gen}} \cup B_{\text{real}}\}$ with transformations.
		\STATE \textcolor{BlueViolet}{\# Step 2: Training main contrastive model}
		\STATE Compute loss $\mathcal{L}_{\text{PPCL}}$ (Eq.(\ref{equ:loss_ppcl})) and optimize $f$ by Eq.(\ref{equ:update_main_and_g}) to minimize $\mathcal{L}_{\text{PPCL}}$.
		\STATE \textcolor{BlueViolet}{\# Step 3: Updating generator}
		\STATE In every $n$ iterations, update $G$ to maximize $\mathcal{L}_{\text{PPCL}}$, update $G_{\text{ema}}$; optimize both $G$ and $D$ by the unconditional GAN objective without using labels.
		\ENDFOR
	\end{algorithmic}
\end{algorithm}
\end{minipage}%
\end{figure}

\subsection{Hard Data Generation}

In this subsection, we first introduce the details of the hard sample generator. The generator $G$ generates synthetic data to augment the training data and the contrastive loss is:
\begin{equation}\label{equ:loss_gen_real}
\mathcal{L}_\text{gen+real} = \sum_{i \in \{ I_{\text{gen}} \cup I_{\text{real}} \} }{\mathcal{L}_i}.
\end{equation}
where $\mathcal{L}_i$ is the contrastive loss of multiviewed sample $i$ defined in Eq.(\ref{equ:cl_loss}). $I_{\text{gen}}$ is the set of indices of generated and then transformed (multiviewed) samples $\{\tilde{x}_{\text{gen}}^{i}\}_{i \in I_{\text{gen}}}$, and $I_{\text{real}}$ is the set of indices of multiviewed real samples.
The generated raw samples $\{x_{\text{gen}}^k\}=\{G(z_k)\}$ are from the generator $G$ by taking a set of vectors $\{z_k\} \sim p(z)$, drawn from a Gaussian distribution $p(z)=\mathcal{N}(0,I)$, as input.
Two transformations are then applied to each $x_{\text{gen}}^k$ to get two views $\tilde{x}_{\text{gen}}^{2k-1}$ and $\tilde{x}_{\text{gen}}^{2k}$ to form mutltviewed samples $\{\tilde{x}_{\text{gen}}^{i}\}_{i \in I_{\text{gen}}}$.

As shown in Fig. \ref{fig:motivation}, simply using a generator to provide additional synthetic data cannot improve contrastive learning.
To generate samples that benefit contrastive learning, we form a Min-Max game to jointly optimize the generator and the contrastive model. 
In this way, the generator dynamically adapts to the training state of the main model and generates hard samples (i.e. high-quality samples from the perspective of training the main model).
The dynamically customized hard samples in each training state of the main model will explore its weakness and encourage it to learn better representations to compete with the generator.
Formally, the joint learning objective is defined as follows.
\begin{equation}\label{equ:minimax}
\min _{\theta} \max _{w} \mathcal{L}_\text{gen+real}.
\end{equation}
where $\theta$ and $w$ are the parameters of the main model and the generator, respectively.
To solve the Min-Max game, a pair of gradient descent and ascent are applied to the main model and the generator to update their parameters, respectively. The details of the update are shown as follows.
\begin{equation}\label{equ:update_main_and_g}
\theta \leftarrow \theta - \eta_{\theta} \frac{\partial \mathcal{L}_\text{gen+real}}{\partial \theta}, \quad 
w \leftarrow w + \eta_{w} \frac{\partial \mathcal{L}_\text{gen+real}}{\partial w}.
\end{equation}
where $\eta_{\theta}$ and $\eta_{w}$ are learning rates for the main model and generator, respectively.

\subsection{Positive Pair Generation without Using Labels}

In addition to generating hard samples, we also propose a new method to generate hard positive pairs. The main idea is that we can use two similar yet different generators $G$ and $G_{\text{ema}}$ to generate two similar but distinct samples as a positive pair, when taking the same latent vector as input.
In joint learning, the hardness of a positive pair is further increased by decreasing their similarity for better CL.

\textbf{Positive pair generation.}
To generate a positive pair, we use a generator $G$ and its slowly-evolving version $G_{\text{ema}}$, which are very similar but different. 
A latent vector $z_i$ is randomly sampled from a Gaussian distribution and serve as the \texttt{pseudo label}. Then $z_i$ is fed to both $G$ and $G_{\text{ema}}$ to generate a pair of raw samples $(x_{2i-1}, x_{2i})$ as a positive pair, which are similar but distinctive. 
\begin{equation}\label{equ:generate_pair}
x_{2i-1} = G(z_i), \quad x_{2i} = G_{\text{ema}}(z_i), \quad z_i \sim p(z).
\end{equation}
To make the positives harder by increasing their difference, in joint learning $G$ is updated with gradients from the main model by Eq.(\ref{equ:update_main_and_g}) while $G_{\text{ema}}$ is not.
On the other hand, to keep the similarity of generated positive pairs, we update $G_{\text{ema}}$ by momentum update following $G$.
Denoting the parameters of $G$ as $w$ and the parameters of $G_{\text{ema}}$ as $w_{\text{ema}}$, $w_{\text{ema}}$ is updated by:
\begin{equation}\label{equ:g_ema_update}
w_{\text{ema}} \leftarrow m w_{\text{ema}} + (1-m)w.
\end{equation}
where $m \in (0,1)$ is a momentum parameter. 

To generate $N$ samples with $\frac{N}{2}$ positive pairs, we sample $\frac{N}{2}$ latent vectors $\{z_{i}\}_{i=1...\frac{N}{2}}$ to generate a batch of $N$ raw samples $B_{\text{gen}}=\{x_k\}_{k=1...N}$ following Eq.(\ref{equ:generate_pair}).
To leverage the high quality of real samples, we further sample $N$ real samples $B_{\text{real}}=\{x_k\}_{k=N+1...2N}$ from the dataset. A raw batch is formed as $B= B_{\text{gen}} \cup B_{\text{real}} = \{x_k\}_{k=1...2N}$. Then two transformations are applied to each $x_k$ to form a multiviewed batch $B_{\text{mv}}=\{ \tilde{x}_{l} \}_{l=1...4N}$ for training as shown in Step 1 of Algorithm~\ref{alg:diffhard}.

\textbf{Partially pseudo-labeled contrastive loss (PPCL).} The contrastive loss in Eq.(\ref{equ:loss_gen_real}) 
only uses two views of a raw sample
as positive pairs. It does not leverage the fact that samples generated by $G$ and $G_{\text{ema}}$ by using the same pseudo label $z$ are actually positive pairs. 
To better cluster the generated positive pairs,
we define a partially pseudo-labeled contrastive loss by using the input latent vectors $z_i$ as pseudo-labels. Each $z_i \ (i=1...\frac{N}{2})$ generates two positives ($x_{2i-1}$, $x_{2i}$) in the raw batch and four positives ($\tilde{x}_{4i-3}$, $\tilde{x}_{4i-2}$, $\tilde{x}_{4i-1}$, $\tilde{x}_{4i}$) in the multiviewed batch,
which are assigned the same pseudo-label for clustering their representations.

Within the multiviewed batch $B_{\text{mv}}$, let $i\in I_{\text{gen}}=\{1...2N\}$ be the indices of generated samples and $i\in I_{\text{real}}=\{2N+1...4N\}$ be the indices of real samples. The PPCL loss is defined as follows.
\begin{equation}\label{equ:loss_ppcl}
\begin{split}
\mathcal{L}_{\text{PPCL}} =& \sum_{i\in I} \frac{-1}{|P(i)|} \sum_{p \in P(i)} \log \frac{\exp{(v_i \cdot v_p / \tau)}}{\sum_{a \in A(i)}{\exp (v_i \cdot v_a / \tau)}}, \\
P(i) =& \begin{cases}
\{ p \in A(i), z_p=z_i \}, & \text{if } i \in I_{\text{gen}}\\
\{j(i)\}, & \text{if } i \in I_{\text{real}}.
\end{cases}
\end{split}
\end{equation}
where $I=I_{\text{gen}} \cup I_{\text{real}}$. $A(i)=I \backslash \{i\}$ is the set of indices of $i$'s positives and negatives, and $P(i)$ is the set of indices of $i$'s positives in the multiviewed batch.
For real samples $i\in I_{\text{real}}$, 
the positive $P(i)=j(i)$ is the index of the other view of $i$ in the multiviewed batch. 
For generated samples $i\in I_{\text{gen}}$, 
the positives $P(i)$ are defined by the pseudo-labels from the input latent vector $z_i$, which includes the indices of multiviewed samples originating from the same $z_i$.

\subsection{Joint Learning}
In this subsection, we illustrate the learning process for the CL model to minimize its loss and the generator to maximize the CL loss.
We rewrite the PPCL loss in Eq.(\ref{equ:loss_ppcl}) as follows.
\begin{equation}\label{equ:loss_ppcl_rewrite}
\mathcal{L}_{\text{PPCL}} = \Sigma_{i\in I} L_i(v_i,\{v_p\}, \{v_a\}),
\end{equation}
where $v_i$ is the representation of anchor $i$, $p \in P(i)$ are positives of the anchor $i$ and $a \in A(i)$ include both positives and negatives as defined in Eq.(\ref{equ:loss_ppcl}).
The representation $v$ is generated as follows.
For clear illustration purposes, we only use one view (i.e., applying $T$ once for one sample) instead of two views.
For synthetic samples, taking a random latent vector $z$ as input, we have a positive pair $(v_j, v_{j+1})$:
\begin{equation}\label{equ:x_from_g}
v_j = f(\tilde{x}_j; \theta), \quad \tilde{x}_j=T(G(z; \omega)).
\end{equation}
\begin{equation}\label{equ:x_from_g_ema}
v_{j+1} = f(\tilde{x}_{j+1}; \theta), \quad \tilde{x}_{j+1}=T(G_\text{ema}(z; \omega_\text{ema})).
\end{equation}
For real samples, we have: 
$v_j = f(\tilde{x}_j; \theta), \tilde{x}_j=T(x_j)$.

The generator $G$ has parameters $\omega$ and the main model $f$ has parameters $\theta$.
By applying the chain rule, the gradient of $\mathcal{L}_{\text{PPCL}}$ w.r.t the parameters $\omega$ of generator G is:
\begin{equation}\frac{\partial \mathcal{L}_{\text{PPCL}}}{\partial \omega} =
\frac{\partial \mathcal{L}_{\text{PPCL}}}{\partial v} \cdot 
\frac{\partial f(\tilde{x}; \theta)}{\partial \tilde{x}} \cdot 
\frac{\partial \tilde{x}}{\partial G(z; \omega)} \cdot
\frac{\partial G(z; \omega)}{\partial \omega},
\end{equation}
where the first item is calculated by Eq.(\ref{equ:loss_ppcl_rewrite}) and other three by Eq.(\ref{equ:x_from_g}) and Eq.(\ref{equ:x_from_g_ema}).
Then the generator is updated by gradient ascent to maximize the CL loss and the main model by gradient descent to minimize the CL loss by Eq.(\ref{equ:update_main_and_g}):

\section{Experimental Results}\label{sec:experiments}

\textbf{Datasets and model architecture.}
We evaluate the proposed approaches on five datasets, including ImageNet-100, CIFAR-10, CIFAR-100 \cite{krizhevsky2009learning}, Fashion-MNIST \cite{xiao2017fashion} and ImageNet-10. 
ImageNet-100 is widely used in contrastive learning \cite{kalantidis2020hard,van2020scan} 
and is a subset of ImageNet \cite{ILSVRC15}. ImageNet-10 is a smaller subset of ImageNet.
We use ResNet-18 as the main model by default unless specified. 
A 2-layer MLP is used to project the output to 128-dimensional representation space \cite{chen2020simple,he2020momentum}. 
We use the generator and discriminator architectures from \cite{brock2018large}.
The batch size is 256 and the main model is trained for 100 epochs on ImageNet-100 for efficient evaluation, 300 epochs on CIFAR-10, CIFAR-100 and ImageNet-10, and 200 epochs on Fashion-MNIST.
The details of training 
and model architectures are in the Appendix.

\textbf{Metrics.}
To evaluate the quality of learned representations, we use two metrics \textit{linear classification} and \textit{transfer learning} widely used for self-supervised learning \cite{chen2020simple}. 
In linear classification, a linear classifier is trained on the frozen encoder, and the test accuracy represents the quality of learned representations. 
We first perform CL by the proposed approaches without labels to learn representations. Then we \textit{fix} the encoder and train a linear classifier on 100\% labeled data on the encoder.
The classifier is trained for 500 epochs with Adam optimizer and learning rate 3e-4.
Transfer learning evaluates the generalization of learned features.
The encoder is learned on the source dataset, then evaluated on the target task. Following \cite{caron2020unsupervised}, we train a linear classifier on the frozen encoder on the target task.

\textbf{Baselines.}
We first compare the performance of our methods with SOTA unsupervised contrastive learning methods \cite{zbontar2021barlow, caron2020unsupervised, grill2020bootstrap, he2020momentum, chen2020improved, chen2020simple}  
to show the effectiveness of our synthetic data generation in improving CL.
Then, we compare our methods with other data generation approaches.
For each mini-batch, the first half of data are the same in different methods and are sampled directly from the training data. The second half of the data are different and are formed by each method as follows.
\textit{SimCLR-DD} is a variant of \textit{SimCLR} by sampling additional real data from the dataset as the second half of mini-batch (i.e. Double Data), serving as a strong baseline. By comparing our approaches with \textit{SimCLR-DD}, we evaluate if the samples generated by our methods benefits CL training more than additional real data.
\textit{CLAE} is the SOTA data generation approach for CL by using pixel-level adversarial perturbations of the real data as the additional data \cite{ho2020contrastive}.
\textit{BigGAN} uses a generator from BigGAN \cite{brock2018large} to generate synthetic training data for \textit{SimCLR} \cite{chen2020simple} without joint learning.
By comparing with \textit{BigGAN}, we evaluate if the hard samples generated by our approaches benefit CL training more than synthetic data generated by a standalone BigGAN.

\subsection{Main Results}

\begin{table}[!htb]
	\centering
    \caption{\textbf{Comparisons on ImageNet-100 linear classification}. All are based on ResNet-50 trained for 200 epochs. }
	\label{tab:exp_more_methods}
	\resizebox{0.8\columnwidth}{!}{
		\begin{tabular}{lccccccccc}
			\toprule
			Method &  Top-1 Acc. \\ \midrule
			SWAV \cite{caron2020unsupervised} & 69.20 &  \\
			BYOL \cite{grill2020bootstrap}  & 75.80 &  \\
			Barlow Twins \cite{zbontar2021barlow} & 77.02 &  \\
			MoCo \cite{he2020momentum} & 76.60 &  \\
			MoCo v2 \cite{chen2020improved} & 78.00 &  \\ \midrule
			SimCLR \cite{chen2020simple} & 75.75 &  \\
            \textbf{SimCLR + Ours}   & \textbf{78.85}  \color{green(pigment)} ({\textbf{+3.1}})  \\ 
			\bottomrule
	\end{tabular}}
\end{table}

\textbf{Comparison with SOTA.}
We first compare the performance of our methods with SOTA unsupervised contrastive learning methods to show the effectiveness of our synthetic data generation in improving CL.
Kindly note that our primary goal is not developing a new unsupervised contrastive learning method to achieve SOTA accuracy. Instead, the goal is to generate effective synthetic data for improving existing CL methods.

The comparison is shown in Table \ref{tab:exp_more_methods}. To be consistent with existing works, we use ResNet-50 as the backbone. First, our data generation methods integrated with SimCLR outperform other methods, while vanilla SimCLR does not.
Second, +3.1\% improvement over the SimCLR baseline is observed, which verifies the effectiveness of our hard sample and hard positive pair generation methods for improving CL.

Next, since our methods are data generation methods, we focus on comparing with other data generation methods, which shows simple GAN-based synthetic generation cannot improve or even degrade CL, while our methods can effectively improve CL.

\begin{table}[!htb]
	\centering
	\caption{\textbf{Linear classification.} 100\% labeled data are used for learning the classifier on the \textit{fixed} encoder and top-1 accuracy is reported.}
	\label{tab:exp_linaer_all_data}
	\setlength\tabcolsep{3.0pt}
	\renewcommand{\arraystretch}{1.0}
	\resizebox{1.0\columnwidth}{!}{
    \begin{tabular}{lcccccccc}
    \toprule
Method                                           & CIFAR-10       & CIFAR-100            & FMNIST           & ImageNet-100            & ImageNet-10           \\ \midrule
SimCLR                     & 90.37          & 63.93         & 92.35          & 67.45          & 83.20          \\
SimCLR-DD                  & \underline{91.37}          & \underline{64.88}         & 92.50          & \underline{70.72}          & 82.20          \\
CLAE                    & 90.13          & 63.25         & 92.36          & 66.40          & 81.20          \\
BigGAN   & 89.90          & 63.82         & \underline{92.64}          & 68.70          & \underline{83.40}          \\
SimCLR+Ours      & \textbf{92.94}          & \textbf{67.41}         & \textbf{93.94}          & \textbf{71.40}          & \textbf{87.40}          \\
    \bottomrule
    \end{tabular}}
\end{table}

\textbf{Linear separability of learned representations.}
We evaluate the proposed approaches by linear evaluation with 100\% data labeled for training the classifier on top of the \textit{fixed} encoder learned with unlabeled data by different approaches. This metric evaluates the linear separability of learned representations \cite{chen2020simple}, and higher accuracy indicates more discriminative and desirable features. 
The proposed approaches significantly outperform the SOTA approaches.
As shown in Table \ref{tab:exp_linaer_all_data}, substantial improvements of 2.57\%, 3.48\%, 1.59\%, 3.95\%, and 4.20\% over the original contrastive learning framework SimCLR are observed on five datasets, respectively.
Interestingly, the proposed approaches even largely outperform SimCLR-DD, which samples $2\times$ real data in each training batch. 
This result shows that the customized hard data by the proposed methods benefits CL more than additional real data.

\begin{table}[!htb]
	\centering
	\caption{\textbf{Linear classification 
	when less training data are available for CL.} Available data are labeled and used for learning the classifier on the \textit{fixed} encoder and top-1 accuracy is reported. Dataset names are abbreviated for conciseness and the percentage means the available training data.}
	\label{tab:exp_linaer_less_data}
	\setlength\tabcolsep{3.0pt}
	\renewcommand{\arraystretch}{1.0}
	\resizebox{1.0\columnwidth}{!}{
    \begin{tabular}{lcccccccccc}
    \toprule
\multirow{2}{*}{Method}                                           & \begin{tabular}[c]{@{}c@{}}C-10\\ (20\%)\end{tabular}       & \begin{tabular}[c]{@{}c@{}}C-10\\ (10\%)\end{tabular}            & \begin{tabular}[c]{@{}c@{}}C-100\\ (20\%)\end{tabular}           & \begin{tabular}[c]{@{}c@{}}C-100\\ (10\%)\end{tabular}            & \begin{tabular}[c]{@{}c@{}}FM\\ (20\%)\end{tabular}            & \begin{tabular}[c]{@{}c@{}}FM\\ (10\%)\end{tabular}      & \begin{tabular}[c]{@{}c@{}}IN-100\\ (20\%)\end{tabular}      & \begin{tabular}[c]{@{}c@{}}IN-10\\ (20\%)\end{tabular}            \\ \midrule
SimCLR                     & 80.19          & 76.05         & 44.13          & 37.46          & 87.92          & 87.26     & 50.09    & 62.60      \\
SimCLR-DD                  & \underline{81.04}          & 76.06         & 44.75          & 37.10          & \underline{89.60}          & 87.49     & \underline{53.38}    & \underline{65.00}      \\
CLAE                    & 79.84          & 74.38         & 45.17          & 37.28           & 88.48          & 87.58     & 49.55    & 59.60      \\
BigGAN   & 80.81          & \underline{78.36}         & \underline{46.94}          & \underline{40.03}          & 88.85          & \underline{87.68}     & 52.49    & 64.20      \\
SimCLR+Ours      & \textbf{85.11}          & \textbf{80.94}         & \textbf{50.24}          & \textbf{43.74}          & \textbf{91.77}          & \textbf{89.79}     & \textbf{55.33}    & \textbf{68.40}      \\
    \bottomrule
    \end{tabular}}
\end{table}

\textbf{Linear separability of learned representations with less training data.}
We evaluate the proposed approaches by linear evaluation when less training data is available for training the encoder.
As shown in Table \ref{tab:exp_linaer_less_data}, the proposed approaches consistently outperform the SOTA approaches by a large margin.
Notably, the proposed approaches outperform or perform on par with the best-performing baselines trained with $2\times$ data, achieving $2\times$ data efficiency.
For example, with 10\% training data on CIFAR-10, the best-performing baseline with 20\% training data achieves 81.04\% accuracy, while the proposed approaches achieve a similar accuracy of 80.94\% by using only 10\% training data.

\begin{table}[!h]
	\centering
	\caption{\textbf{Transfer learning to downstream tasks.} Top-1 accuracy of linear classification is reported.}
	\label{tab:exp_transfer_full_data}
	\setlength\tabcolsep{3.0pt}
	\renewcommand{\arraystretch}{1.0}
	\resizebox{0.9\columnwidth}{!}{
    \begin{tabular}{lccccccccc}
    \toprule
Source     & C-10                     & C-100   & \multicolumn{2}{c}{IN-100}  & \multicolumn{2}{c}{IN-10}              \\ \cmidrule(lr){1-1} \cmidrule(lr){2-2} \cmidrule(lr){3-3} \cmidrule(lr){4-5} \cmidrule(lr){6-7}
Target     & C-100                    & C-10    & C-10  & C-100  & C-10   & C-100              \\ \midrule
SimCLR                    & 51.63          & 78.87          & 80.84          & 56.06         & 73.91          & 44.88           \\
SimCLR-DD                 & 52.22          & 78.53          & 81.10          & 56.03         & 73.77          & 46.53           \\
CLAE                   & \underline{52.47}          & 78.10          & 80.80          & \underline{57.08}         & \underline{74.07}          & \underline{47.11}           \\
BigGAN  & 52.16          & \underline{78.98}          & \underline{81.13}          & 56.59         & 73.62          & 44.55           \\
SimCLR+Ours                                        & \textbf{56.38}          & \textbf{82.24}          & \textbf{83.25}          & \textbf{58.61}         & \textbf{75.38}          & \textbf{49.03}           \\
    \bottomrule
    \end{tabular}}
\end{table}

\textbf{Transfer learning.}
We evaluate the generalization of learned representations by transferring to downstream tasks.
In Table \ref{tab:exp_transfer_full_data}, the encoder is trained on the source dataset and transferred to target tasks, and we report the linear classification performance. Our approaches outperform the baselines on four source datasets with various target tasks. 

\begin{table}[!htb]
	\centering
	\caption{\textbf{Transfer learning to downstream tasks with less training data for learning the encoder on the source datasets.} Top-1 accuracy of linear classification on top of a \textit{fixed} encoder is reported. }
	\label{tab:exp_transfer_less_data}
	\setlength\tabcolsep{3.0pt}
	\renewcommand{\arraystretch}{1.0}
	\resizebox{1.0\columnwidth}{!}{
    \begin{tabular}{lccccccccc}
    \toprule
Source     & \begin{tabular}[c]{@{}c@{}}C-10\\ (20\%)\end{tabular}                     & \begin{tabular}[c]{@{}c@{}}C-10\\ (10\%)\end{tabular}   & \begin{tabular}[c]{@{}c@{}}C-100\\ (20\%)\end{tabular}  & \begin{tabular}[c]{@{}c@{}}C-100\\ (10\%)\end{tabular}  & \multicolumn{2}{c}{\begin{tabular}[c]{@{}c@{}}IN-100\\ (20\%)\end{tabular}}  & \multicolumn{2}{c}{\begin{tabular}[c]{@{}c@{}}IN-10\\ (20\%)\end{tabular}}  &          \\ \cmidrule(lr){1-1} \cmidrule(lr){2-3}  \cmidrule(lr){4-5} \cmidrule(lr){6-7} \cmidrule(lr){8-9}
Target     & \multicolumn{2}{c}{C-100}                    & \multicolumn{2}{c}{C-10}    &  C-10  &  C-100  &  C-10  & C-100               \\ \midrule
SimCLR                    & 46.88          & 44.98          & 74.33          & 71.79    & 76.97 & 50.68 & 69.86 & 42.56     \\
SimCLR-DD                 & 48.45          & 46.74          & 73.94          & 72.49    & 77.13 & 52.46 & \underline{71.20} & 43.82     \\
CLAE                   & 46.93          & 46.11          & 73.47          & 71.48    & \underline{77.79} & \underline{52.63} & 69.98 & 44.19     \\
BigGAN   & \underline{48.54}          & \underline{47.39}          & \underline{75.15}          & \underline{73.43} & 77.54 & 52.28 & 70.44 & \underline{44.74}        \\
SimCLR+Ours                                        & \textbf{52.51}          & \textbf{50.08}          & \textbf{77.26}          & \textbf{75.10} & \textbf{80.10} & \textbf{55.76} & \textbf{73.08} & \textbf{46.46}         \\
    \bottomrule
    \end{tabular}}
\end{table}

\textbf{Transfer learning with less training data on the source dataset.}
We further evaluate the transfer learning performance when less training data is available for learning the encoder on the source datasets.
As shown in Table \ref{tab:exp_transfer_less_data}, the proposed approaches consistently outperform the SOTA baselines. 
Notably, with only 20\% training data (Table \ref{tab:exp_transfer_less_data}) of the source dataset CIFAR-10 and transferring to CIFAR-100, the proposed approaches outperform the best-performing baseline with 100\% training data (Table \ref{tab:exp_transfer_full_data}) of CIFAR-10 (52.51\% vs. 52.47\%), achieving $5\times$ data efficiency.
Besides, with only 10\% training data on CIFAR-10, the proposed approaches outperform the best-performing baseline with 20\% training data by a large margin (50.08\% vs. 48.54\%).

\subsection{Ablations}

\begin{figure}[!htb]
	\centering
	\begin{subfigure}[b]{0.49\linewidth}
		\includegraphics[width=\linewidth]{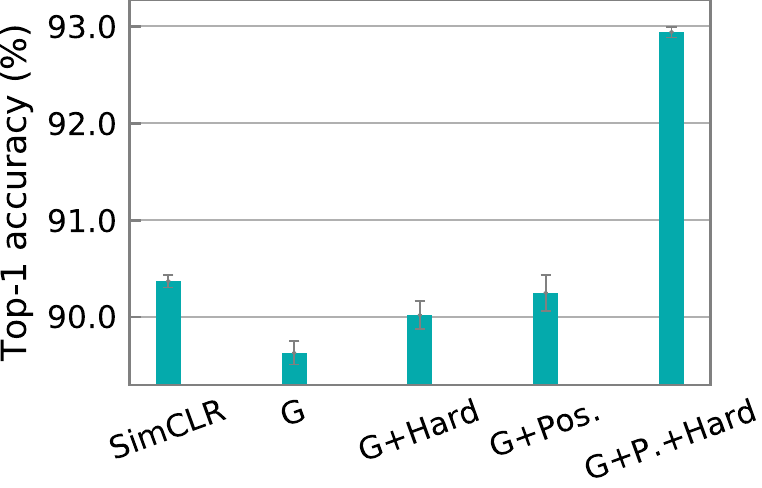}
		\caption{CIFAR-10.}
		\label{fig:}
	\end{subfigure}
	\begin{subfigure}[b]{0.49\linewidth}
		\includegraphics[width=\linewidth]{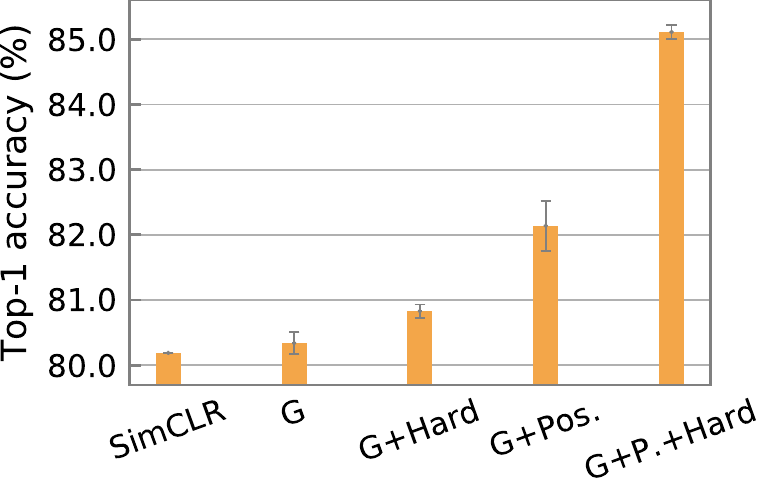}
		\caption{CIFAR-10 (20\% tr. data).}
		\label{fig:}
	\end{subfigure}
	\caption{
		\textbf{Effectiveness of hard samples, positive pairs, and hard positive pairs.}
		Enabling the components in the proposed approaches one by one accumulatively improves the performance. \textit{G} is using a standalone generator, \textit{G+Hard} is jointly optimizing \textit{G} and the main model to generate hard samples, \textit{G+Pos.} is \textit{G} with positive pair generation, and \textit{G+P.+Hard} is the proposed approach with all the components enabled. Top-1 accuracy of linear classification is reported. Error bars are the standard deviations across three independent runs.}
	\label{fig:exp_ablation_positive_cifar_100_020}
\end{figure}

\begin{figure}[!htb]
    \centering
	\includegraphics[width=1.0\columnwidth]{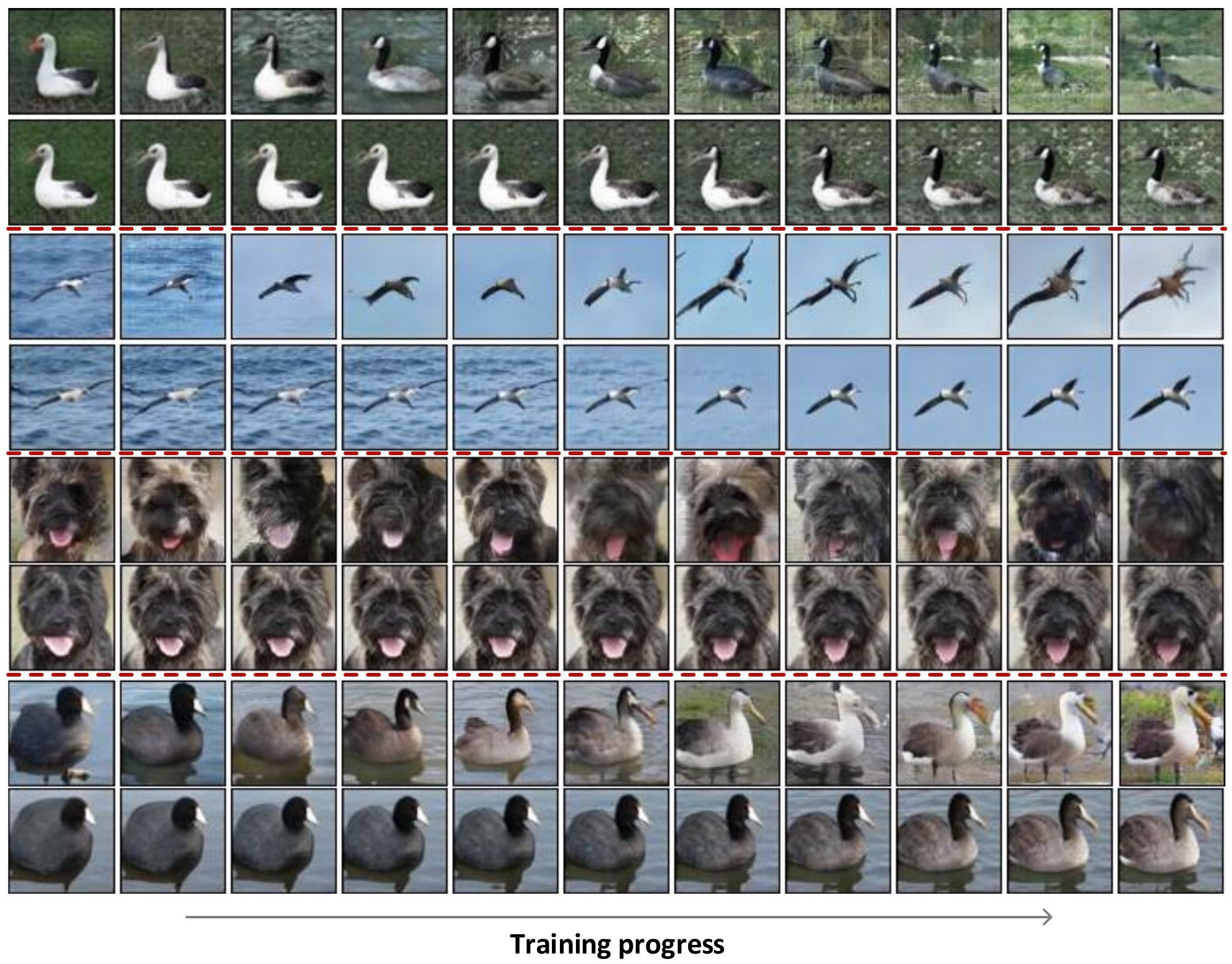}
	\caption{Progressively harder positive pairs during training. }
	\label{fig:exp_visual_hard_positive}
\end{figure}

\textbf{Effectiveness of hard sample generation.}
We perform ablation studies to evaluate the effectiveness of hard samples (\textit{G+Hard}), positive pairs (\textit{G+Pos.}), and hard positive pairs (\textit{G+Pos.+Hard}). The results of linear classification are shown in Fig. \ref{fig:exp_ablation_positive_cifar_100_020}. 
On CIFAR-10, using a simple generator degrades the performance of CL compared with the original SimCLR due to the low-quality samples from the generator.
Using the proposed hard samples and positive pairs recover the accuracy by 0.39\% and 0.62\%, respectively, while using hard positive pairs outperforms SimCLR by 2.57\% (92.94\% vs. 90.37\%).
Similar results are observed on CIFAR-10 with 20\% training data. Using a separate generator only improves the accuracy by 0.15\%, while enabling all the proposed components significantly improves the accuracy by 4.92\%.

\textbf{Evolution of positive pairs.}
The dynamically harder positive pairs in the training progress are shown in Fig. \ref{fig:exp_visual_hard_positive}.
Every two adjacent rows show the evolution of positive pairs on ImageNet-10.
With growing knowledge of the main model, positive pairs become progressively harder, while being similar objects. Learning from harder positive pairs improves the quality of learned representations.

\section{Conclusion}\label{sec:conclusion_limitations}

This paper presents a data generation framework for unsupervised visual representation learning. 
A hard data generator is jointly optimized with the main model to 
customize hard samples for better contrastive learning.
To further generate hard positive pairs without using labels, a pair of generators is proposed to generate similar but distinct samples. 
Experimental results 
show superior accuracy and data efficiency of the proposed data generation methods applied to contrastive learning.

\section{Acknowledgments}
This work was supported in part by NSF CNS-2122320, CNS-212220, CNS-1822099, and CNS-2007302.

\bibliography{ref}

\end{document}